\documentclass[sigconf]{acmart}

\author{Xiaoming Zhang}
\authornote{These authors contributed equally to this research.}
\orcid{0009-0005-0024-224X}
\affiliation{%
   \institution{Northeastern University}
  \city{Shenyang}
  \country{China}  }
\email{zxm2282588541@gmail.com}

\author{Ming Wang}
\authornotemark[1]
\orcid{0000-0001-8406-5677}
\affiliation{%
   \institution{Northeastern University}
  \city{Shenyang}
  \country{China}  }
\email{sci.m.wang@gmail.com}

\author{Xiaocui Yang}
\authornotemark[1]
\affiliation{%
   \institution{Northeastern University}
  \city{Shenyang}
  \country{China}  }
\email{neu.yangxiaocui@gmail.com}

\author{Daling Wang}
\authornote{Corresponding author}
\affiliation{%
   \institution{Northeastern University}
  \city{Shenyang}
  \country{China}  }
\email{wangdaling@cse.neu.edu.cn}

\author{Shi Feng}
\affiliation{%
   \institution{Northeastern University}
  \city{Shenyang}
  \country{China}  }
\email{fengshi@cse.neu.edu.cng}

\author{Yifei Zhang}
\affiliation{%
   \institution{Northeastern University}
  \city{Shenyang}
  \country{China}  }
\email{zhangyifei@cse.neu.edu.cn}
\usepackage{bm}
\usepackage{adjustbox}
\usepackage{multirow}
\usepackage[linesnumbered,ruled,vlined]{algorithm2e}
\usepackage{algpseudocode}

\usepackage{subfigure}
\AtBeginDocument{%
  }

\setcopyright{acmlicensed}
\copyrightyear{2018}
\acmYear{2018}
\acmDOI{XXXXXXX.XXXXXXX}

\acmConference[Conference acronym]{}{2024}


\begin{document}

\title{Hierarchical Retrieval-Augmented Generation Model with Rethink for Multi-hop Question Answering}

\renewcommand{\shortauthors}{Trovato et al.}
\begin{abstract}
    Multi-hop Question Answering (QA) necessitates complex reasoning by integrating multiple pieces of information to resolve intricate questions. However, existing QA systems encounter challenges such as outdated information, context window length limitations, and an accuracy-quantity trade-off. To address these issues, we propose a novel framework, the Hierarchical Retrieval-Augmented Generation Model with Rethink (HiRAG), comprising Decomposer, Definer, Retriever, Filter, and Summarizer five key modules. We introduce a new hierarchical retrieval strategy that incorporates both sparse retrieval at the document level and dense retrieval at the chunk level, effectively integrating their strengths. Additionally, we propose a single-candidate retrieval method to mitigate the limitations of multi-candidate retrieval. We also construct two new corpora, Indexed Wikicorpus and Profile Wikicorpus, to address the issues of outdated and insufficient knowledge.
    Our experimental results on four datasets demonstrate that HiRAG outperforms state-of-the-art models across most metrics, and our Indexed Wikicorpus is effective. The code for HiRAG is available at \url{https://github.com/2282588541a/HiRAG}.
\end{abstract}

\ccsdesc[500]{Information systems~Combination, fusion and federated search}

\keywords{Large Language Models, Hierarchical Retrieval-Augmented Generation, Indexed Wikicorpus Corpus, Profile Corpus}


\maketitle

\section{INTRODUCTION}
Multi-hop Question Answering (QA) involves complex reasoning by integrating multiple pieces of information to resolve intricate questions \cite{mavi2022survey, yang-etal-2018-hotpotqa, ho2020constructing}. Unlike single-hop QA, where answers are readily available, complex questions require decomposing the original query into a series of targeted sub-questions. The knowledge required to answer each sub-question varies, drawing from both internal knowledge encoded in Large Language Models (LLMs) and external knowledge retrieved from local knowledge bases, such as Wikipedia \cite{karpukhin2020dense}, or open search engines like Google \cite{xu2024search, wu2024faithful}. As internal knowledge is derived from large-scale data and extensive pre-training, making it challenging to modify, we focus primarily on updating, mining, and effectively leveraging the retrieved knowledge to enhance QA performance.

\begin{figure}[t!]
    \centering
    \subfigure[Multi-hop QA.]{
          \label{Fig1.sub.Multi-hop QA}
          \includegraphics[scale = 0.28]{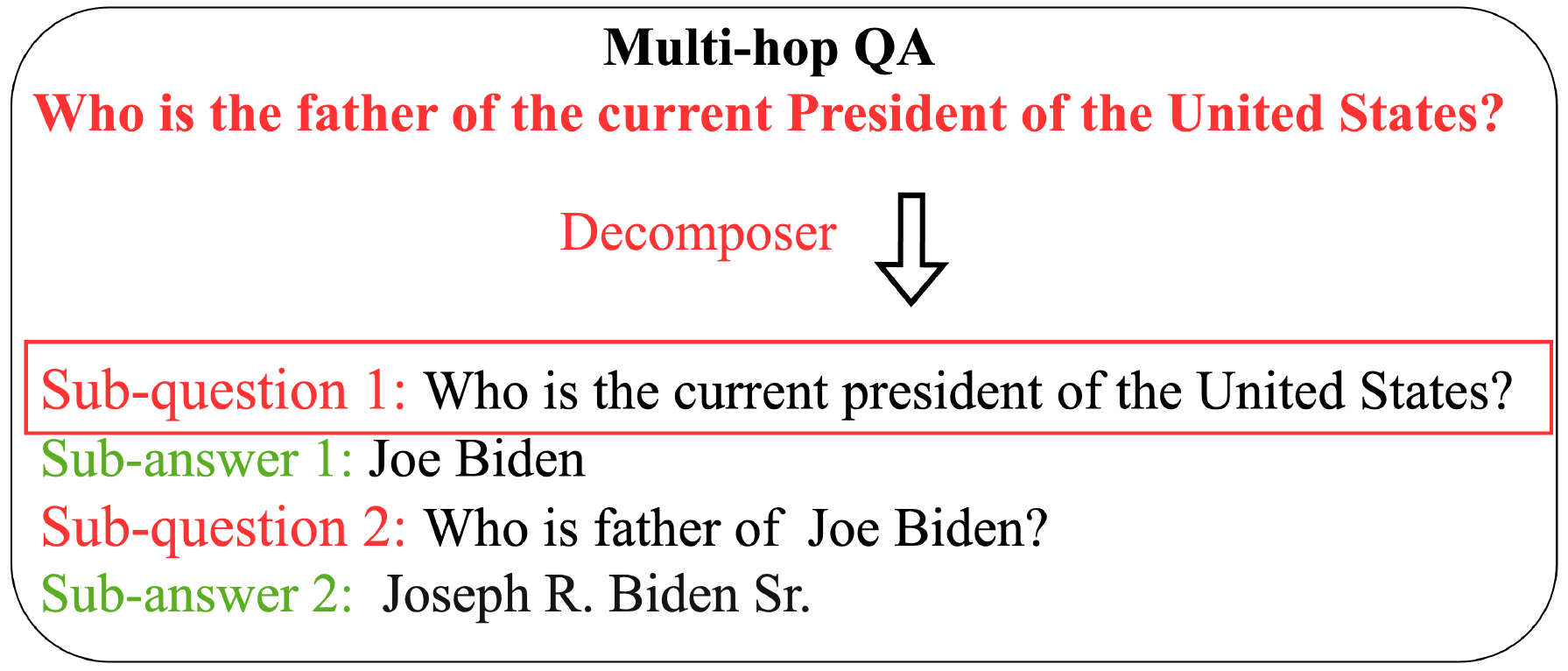}
        }
    \subfigure[Outdated knowledge.]{
          \label{Fig1.sub.a.outdated_knowledge}
          \includegraphics[scale = 0.3]{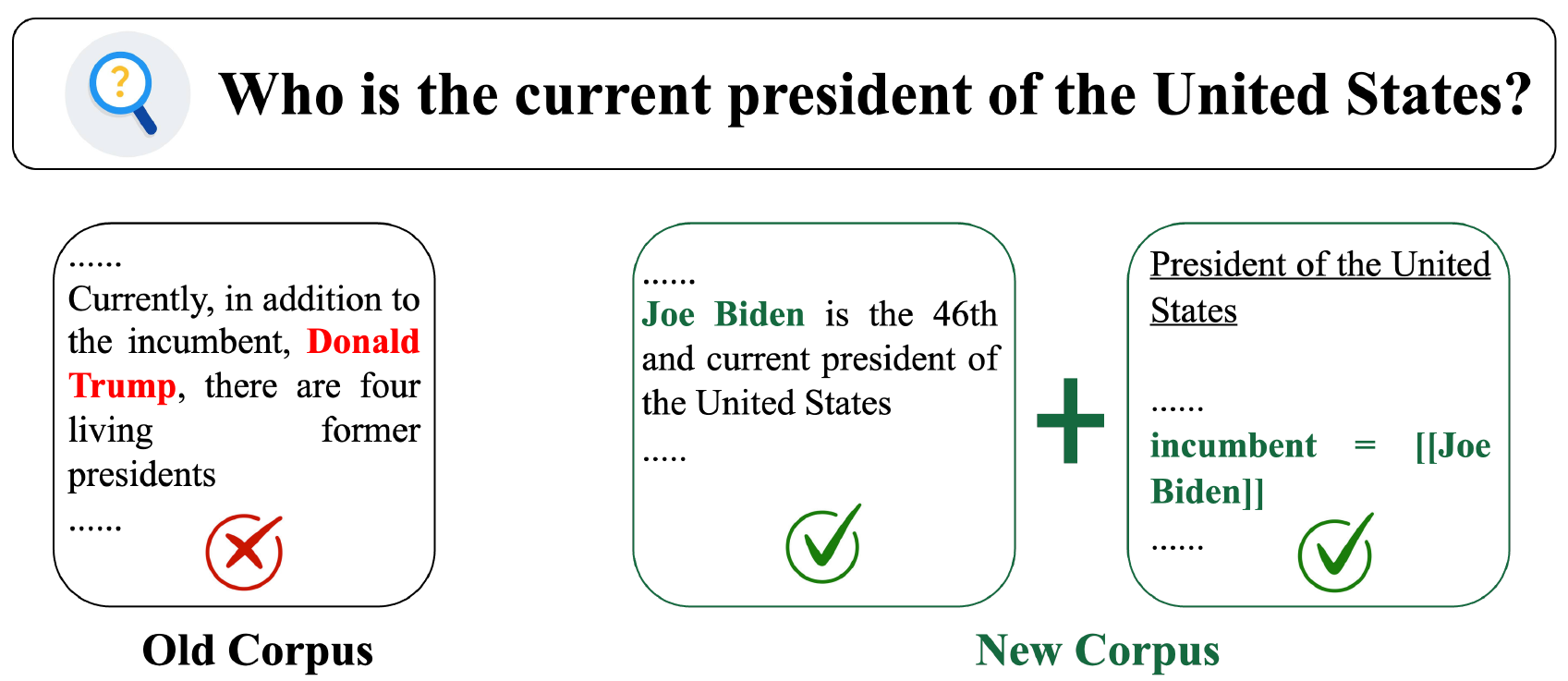}
        }
   \subfigure[Insufficient knowledge.]{
          \label{Fig1.sub.b.insufficient_knowledge}
          \includegraphics[scale = 0.3]{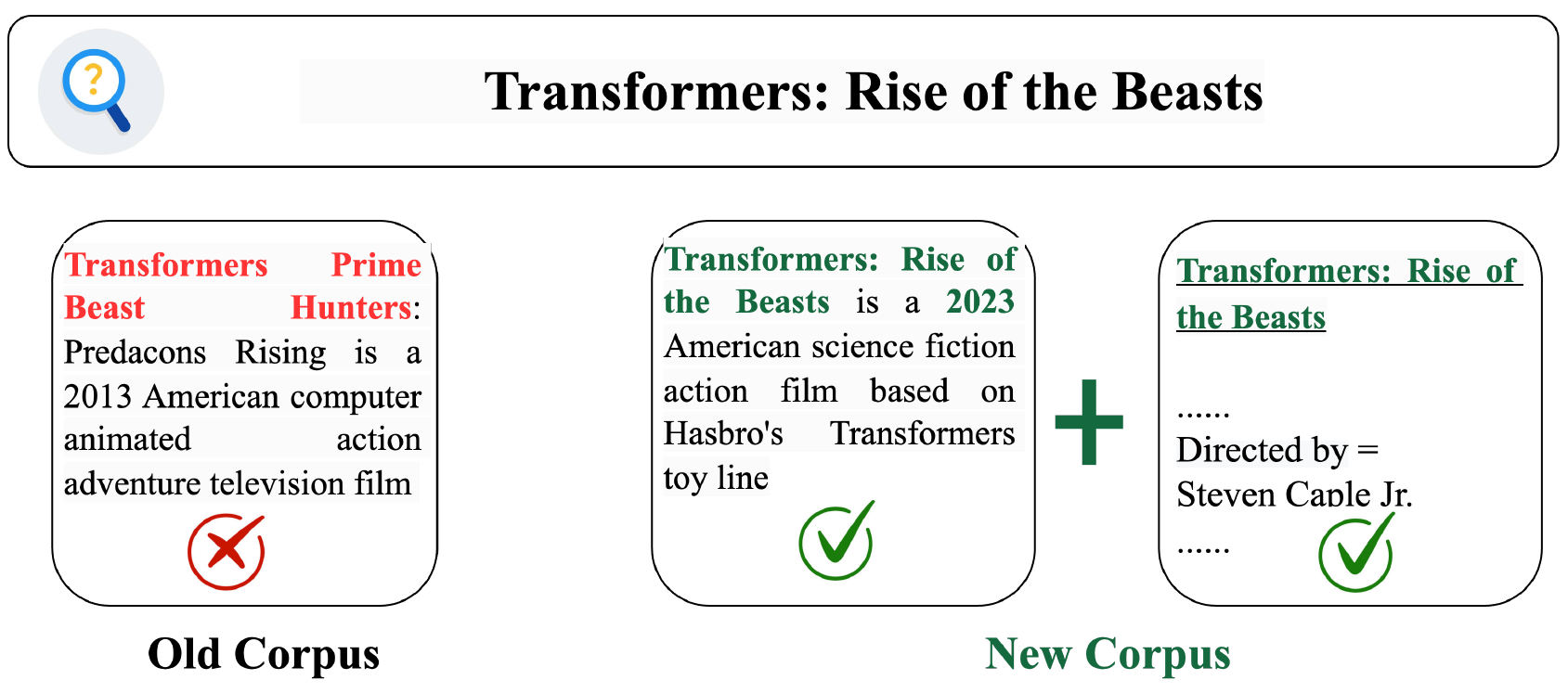}
        }
  \caption{Outdated knowledge and insufficient knowledge existing in the old corpus.
  }
  \label{Fig1.old corpus vs. new corpus}
\end{figure}

\begin{figure*}[t!]
    \centering
    \subfigure[Multiple candidate chunks.]{
          \label{Fig2.sub.a.multiple_chunks}
          \includegraphics[scale = 0.28]{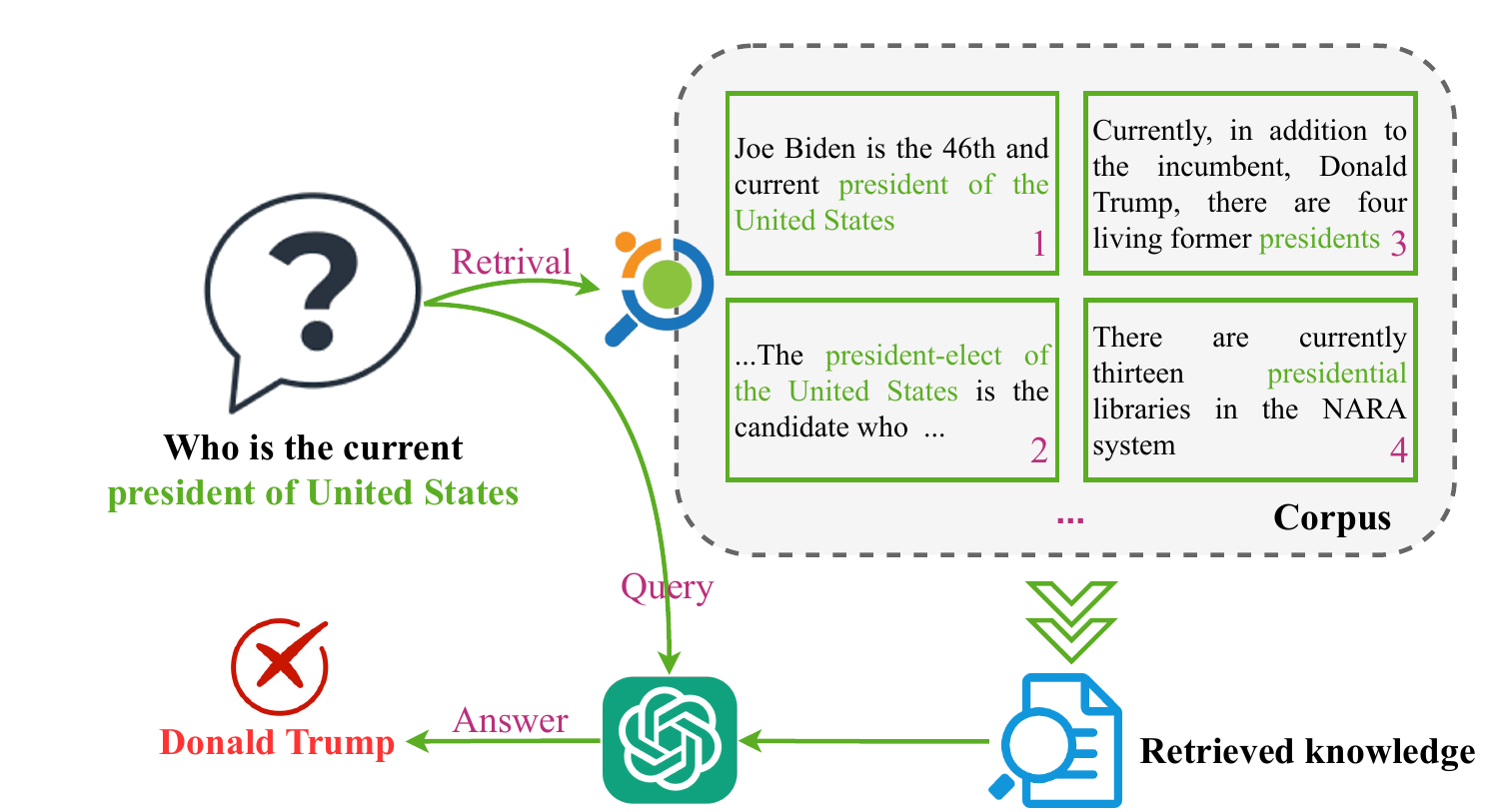}
          \vspace{-0.9em}
        }
   \subfigure[Single candidate chunk.]{
          \label{Fig2.sub.b.single_chunk}
          \includegraphics[scale = 0.28]{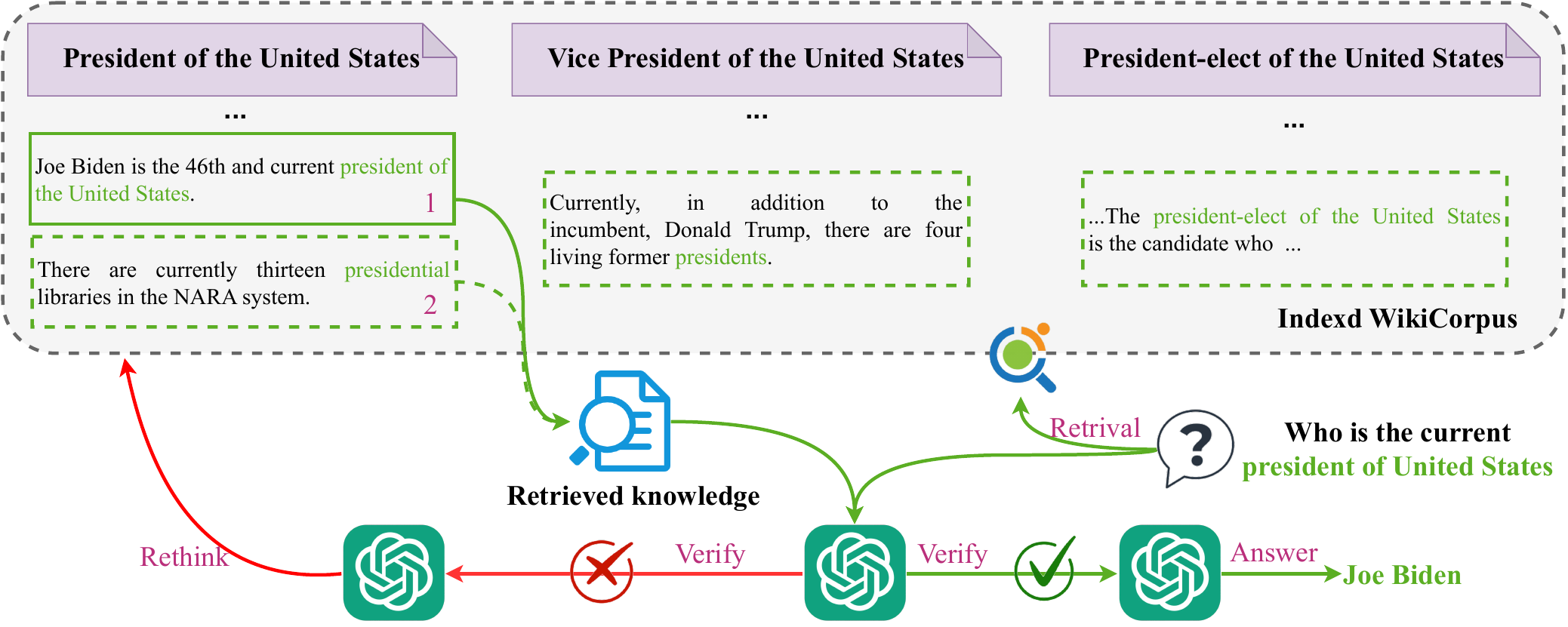}
          \vspace{-0.9em}
        }
  \caption{Multiple candidate chunks vs. single candidate chunk. The knowledge within the solid box represents the retrieved knowledge. In subfigure \ref{Fig2.sub.b.single_chunk}, the knowledge within the dotted box will be considered as newly retrieved knowledge only if the answer to the sub-question is incorrect.
  }
  \label{Fig2.different chunks}
\end{figure*}

The selection and integration of knowledge retrieved from multiple sources pose various challenges. Several key issues, such as outdated information, context window length limitations, and accuracy-quantity trade-off issues, significantly impact the performance of multi-hop QA systems. Firstly, new multi-hop QA systems often rely on outdated and insufficient knowledge within localized knowledge bases. As illustrated in Figure \ref{Fig1.sub.a.outdated_knowledge}, a QA system searches an outdated corpus \cite{karpukhin2020dense} to answer the query \textit{Who is the current president of the United States?}'', resulting in the incorrect answer `Donald Trump'. Similarly, the outdated corpus cannot provide real-time updates for newly released news, movies, books, etc. For instance, it does not include the 2023 movie \textit{Transformers: Rise of the Beasts}'', as shown in Figure \ref{Fig1.sub.b.insufficient_knowledge}. Although outdated information can be mitigated through online retrieval, excessive retrieval introduces efficiency and cost issues. To address this, we first update the older corpus and construct an entity-centric, more comprehensive corpus, \textbf{Indexed Wikicorpus}, based on Wikipedia, to reduce the reliance on over-searching. Additionally, to quickly understand the basic information of entities, we create a corpus containing some basic information of entities, \textbf{Profile Wikicorpus}.

Building on previous studies \cite{press2022measuring, yao2022react, trivedi2022interleaving}, we decompose the original question into multi-hop questions, as illustrated in Figure \ref{Fig1.sub.Multi-hop QA}. Each sub-question is answered based on retrieved knowledge, and the answers are then integrated using the Chain-of-Thought (CoT) approach to derive the final answer.
In local retrieval, most studies, such as FLARE \cite{jiang2023active} and MetaRAG \cite{zhou2024metacognitive}, employ a retrieval method that involves searching the corpus in chunks and returning the top $n$ most similar candidate chunks as retrieved knowledge. We refer to this method as \textit{multi-candidate retrieval}, as depicted in Figure \ref{Fig2.sub.a.multiple_chunks}. Intuitively, increasing the value of $n$ acquires more knowledge, thereby improving the probability of obtaining the correct answer to the query.
However, these approaches face two significant challenges. Firstly, the \textbf{context window limit} imposes an upper limit on the value of $n$, preventing the unlimited expansion of the retrieved content due to the model's context window size constraints. Secondly, the \textbf{accuracy-number tradeoff} arises, where increasing $n$ also increases the amount of potentially irrelevant or redundant information in the retrieved content. This noisy data may not only mislead the LLM and exacerbate its hallucination problem but also distract its attention, causing it to miss relevant information.

To address the challenges, we propose a novel framework called the \textbf{Hi}erarchical \textbf{R}etrieval-\textbf{A}ugmented \textbf{G}eneration Model with Rethink (\textbf{HiRAG}), as illustrated in Figure \ref{fig3:The framework of model}. HiRAG comprises five key modules, i.e., Decomposer, Definer, Retriever, Filter, and Summarizer. The Filter module further incorporates two submodules, including Verify and Rethink. Previous methods typically rely on either sparse retrieval \cite{svore2009machine}, which focuses on lexical matching, or dense retrieval \cite{zhao2024dense}, which leverages latent embeddings for semantic matching. While some approaches combine the results of these two retrieval methods at the output level \cite{arabzadeh2021predicting, luan2021sparse, karpukhin2020dense}, they do not deeply integrate the strengths of each retrieval method within the retrieval process itself.
We propose the Retrieval module which incorporates a new hierarchical retrieval strategy, performing multi-level retrieval from the document level to the chunk level. Initially, we employ sparse retrieval at the document level to extract key information, such as entity names, through lexical matching. Subsequently, dense retrieval is used to retrieve specific information at the chunk level, leveraging semantic matching.
We also introduce a novel retrieval method, \textbf{\textit{single-candidate retrieval}}, as shown in Figure \ref{Fig2.sub.b.single_chunk}. This method returns only the most similar chunk as the retrieved knowledge for each decomposed sub-question. However, selecting just one candidate does not guarantee the correctness of the answer. To mitigate this, we apply the Filter module to assess the answer based on the retrieved knowledge. If the answer to a sub-question is found to be incorrect, we utilize the Rethink module, in conjunction with the Retrieval module, to select another chunk as the retrieved knowledge to answer the question.
The rethinking process is iterated until the Filter module confirms the answer as correct, thereby yielding the solution to the current sub-question. This cycle of decomposition, retrieval, and answering is repeated for subsequent sub-questions until the Definer module determines that the current question, accompanied by its sub-answers, is answerable. At this point, the Summarizer module is invoked to generate the final answer to the current question.

We conduct experiments on four datasets, including HotPotQA \cite{yang-etal-2018-hotpotqa}, 2WikiMultihopQA \cite{ho2020constructing}, MuSiQue \cite{10.1162/tacl_a_00475}, and Bamboogle \cite{press2022measuring}, for multi-hop question answering tasks. The experimental results demonstrate that HiRAG outperforms state-of-the-art models across all metrics on three datasets while showing advancements in several metrics on the remaining dataset, and our Indexed Wikicorpus is effective.
Our main contributions are summarized as follows:

\begin{itemize}
\item To address the limitations of outdated and insufficient knowledge in existing corpora, we construct the Indexed Wikicorpus, which is organized by entity names. Subsequent experiments demonstrate that this corpus is more comprehensive and logically structured. Additionally, we propose the Profile Wikicorpus, which extracts auxiliary information on key entities to further provide effective knowledge.
\item We propose a novel retrieval-augmented generation framework, \textbf{Hi}erarchical \textbf{R}etrieval-\textbf{A}ugmented \textbf{G}eneration Model with Rethink (\textbf{HiRAG}), which not only introduces a hierarchical retrieval method to integrate the advantages of different retrieval technologies but also uses single-candidate retrieval to solve the problems currently encountered in multi-candidate retrieval.
\item We conduct comprehensive experiments on four datasets and verify the effectiveness of different components in HiRAG. The experimental results demonstrate that HiRAG outperforms state-of-the-art models on most metrics and confirms the effectiveness of the Indexed Wikicorpus.
\end{itemize}

\section{RELATED WORK}
Multi-hop QA is the task of answering natural language questions that involve extracting and combining multiple pieces of information and doing multiple steps of reasoning \cite{mavi2024multi}. It can be divided into two steps, a retrieval (IR) step to extract all relevant context from the corpus, and a reading comprehension (MRC) step to find the answer from the reading result context. 
In the era of LLM, research \cite{min2019compositional,du2017learning} finds that on the one hand, LLM can be used to complete the decomposition of the problem and achieve more fine-grained retrieval in the IR step, and on the other hand, Retrieval-Augmented Generation technology can be used to achieve the integration of retrieval information in the MRC step.

\subsection{Retrieval-Augmented Generation}
Retrieval-Augmented Generation (RAG) is a current research hotspot for Multi-hop QA tasks \cite{lewis2020retrieval}. It can integrate the internal knowledge of the model with the external knowledge retrieved. LLM can retrieve external content through RAG to expand their knowledge base, thereby improving their ability to generate accurate and contextually relevant responses.
Historically, various studies attempt to adapt the use of generative models to improve their performance. For instance, REPLUG \cite{shi2023replug} uses different retrieved content to generate corresponding answers and then combine them. Self-Rag \cite{asai2023self} fine-tunes a generation model to simultaneously produce answers along with relevance, support, and usefulness scores. Concurrently, several methods for multi-hop QA emphasize the content and timing of retrieval. Self-Ask \cite{press2022measuring} lets the model generate sub-questions and queries, and continuously alternate between retrieval and generation. PROMPTAGATOR \cite{dai2022promptagator}, Take a step back \cite{zheng2023take} focus on abstracting high-level concepts and utilizing LLMs for prompt-based query generation. Additionally, the confidence-based method, FLARE \cite{jiang2023active}, generates queries using low-confidence tokens. However, most studies directly feed the retrieval content into the generation model, ignoring the evaluation and processing of the retrieval content. Unlike them, HiRAG highlights the importance of verifying retrieved content and adjusts the retriever to enhance the relevance of results when the quality of the retrieved information is subpar.


\subsection{Chain-of-Thought (CoT)}
In the task of Multi-hop QA in addition to using retrieval-enhanced generation to obtain knowledge, CoT is also needed to improve the logic.
CoT can significantly enhance the reasoning capabilities of models \cite{chu2023survey}. For instance, iterative Context-Aware Prompter \cite{wang-etal-2022-iteratively} employs an iterative approach to knowledge acquisition from the model to accomplish reasoning tasks. Similarly, LEAST-TO-MOST \cite{zhou2022least} decomposes complex problems into a series of sub-problems, addressing each step methodically. Inspired by the concept of self-consistency \cite{wang2022self}, MCR \cite{yoran2023answering} extends the application of self-consistency beyond final results to include intermediate steps, thereby enhancing the overall accuracy of reasoning. Following \cite{zhou2022least}, we design a comprehensive prompt to let the model decompose the question into multiple sub-questions and finally integrate the multiple sub-answers into the final answer through CoT. The main difference between our work and previous work is that we do not decompose the problem all at once, but proceed in a loop, generating only one sub-problem in each round; at the same time, we design a matching Definer to realize the judgment of whether the problem can be solved and exit the loop.

\section{INDEXED WIKICORPUS AND PROFILE WIKICORPUS}
\begin{figure}[t]
    \centering
    \includegraphics[scale = 0.3]{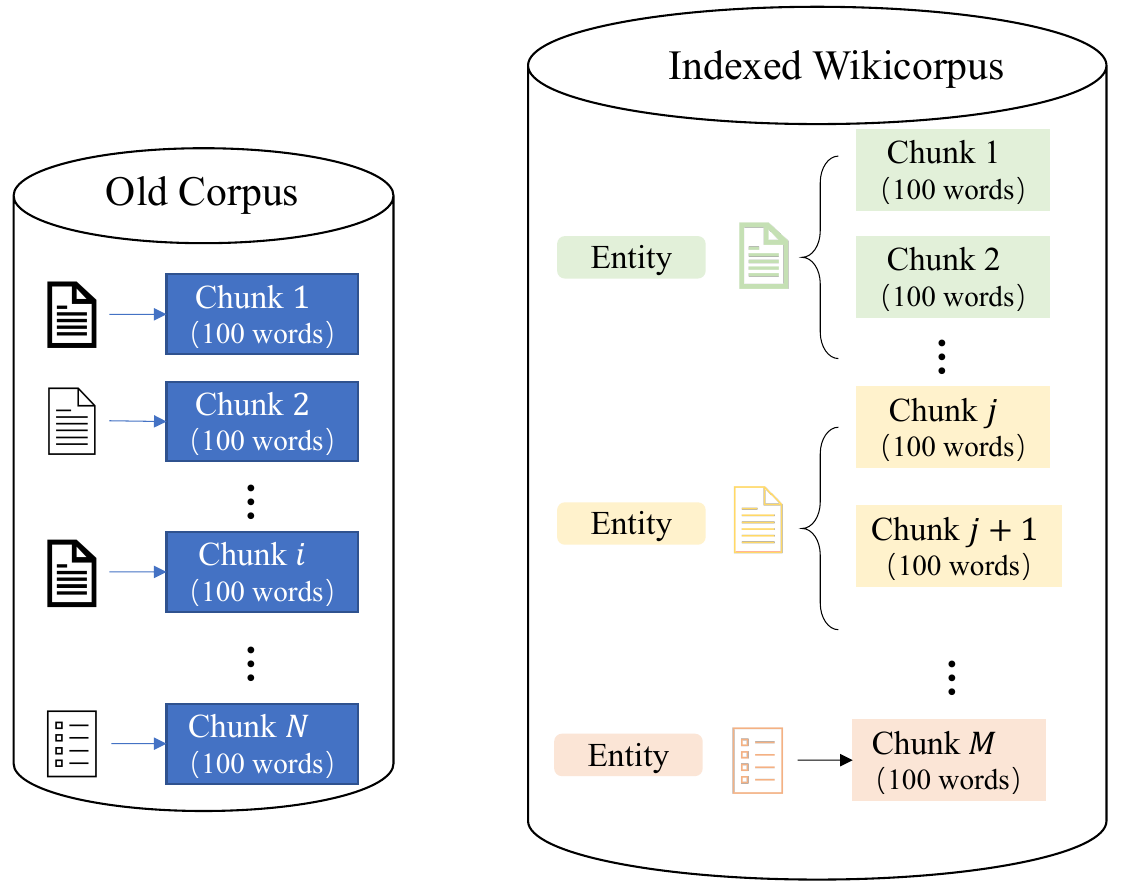}
    \vspace{-1em}
  \caption{Old corpus vs. Indexed Wikicorpus. In the new corpus, we specialize in constructing a document for each entity and then dividing it into chunks. }
  \label{fig:corpus_difference}
\end{figure}

\begin{table}[t]

\renewcommand\arraystretch{1}
\caption{
     Old Corpus vs. Indexed Wikicorpus.
    }
\vspace{-1em}
\begin{adjustbox}{width=0.45\textwidth}
\begin{tabular}{ccc}
 
\toprule[1pt]
\textbf{Corpus}     & \textbf{Number of entities} & \textbf{Number of words} \\ \midrule[1pt]
Old corpus (from DPR) & 3232908         & 2101532400   \\
Indexed Wikicorpus & \textbf{6416724} & \textbf{2642615682}  \\ \bottomrule[1pt]
\end{tabular}
\end{adjustbox}
\label{tab:corpus_diff}
\end{table}

\begin{figure*}[t!]
    \centering
    \includegraphics[width=0.95\textwidth]{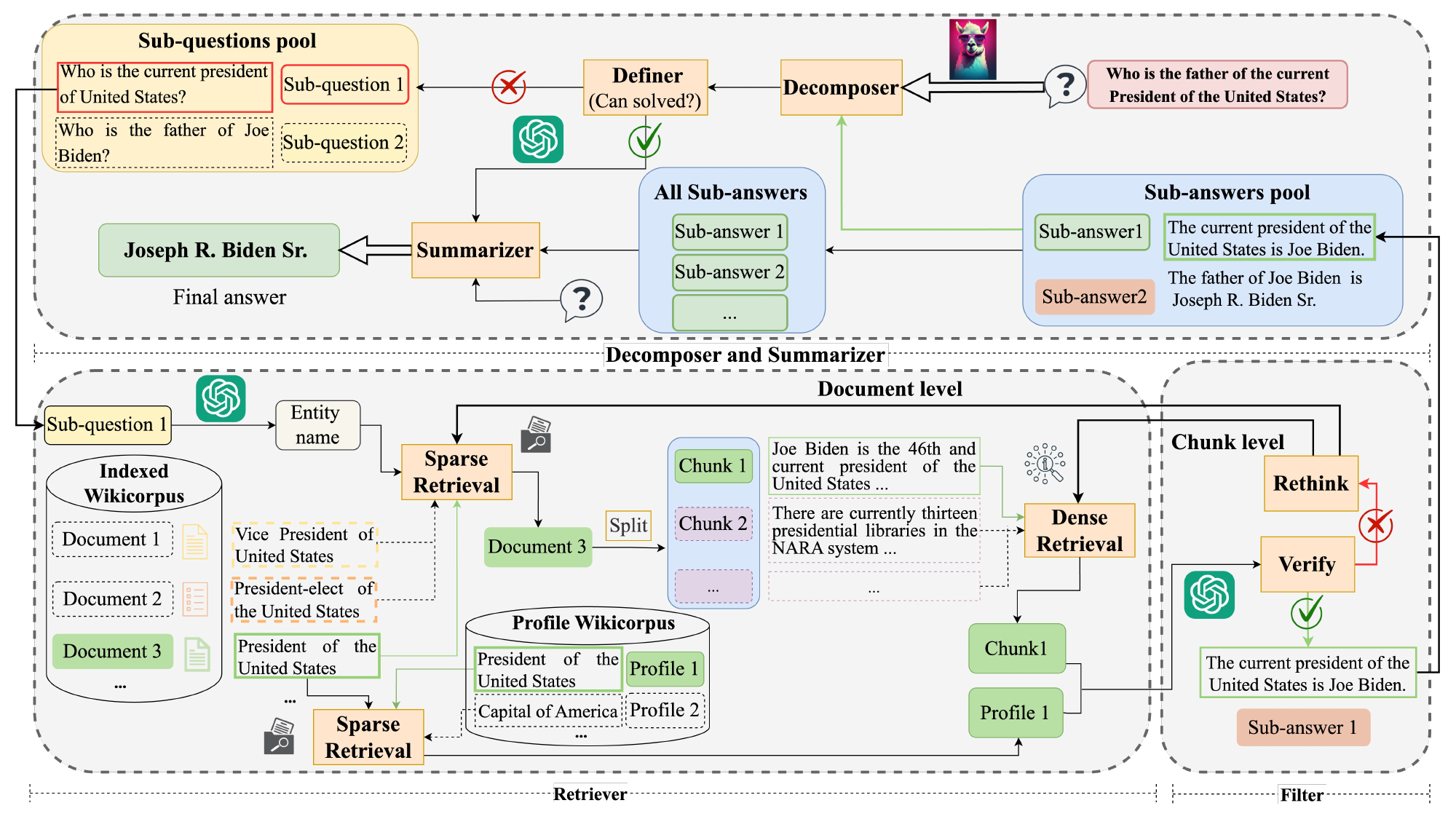}
    \vspace{-1em}
  \caption{The framework of the Hierarchical Retrieval-Augmented Generation Moel with Rethink (\textbf{HiRAG}), including Decomposer module, Definer module, Retriever module, Filter module, and Summarizer module. The process begins with the Decomposer module, which breaks down the original question into several sub-questions. Each sub-question is then forwarded to the Retriever module to retrieve pertinent knowledge. \textit{For illustration, we consider the handling of the first sub-question as an example}. The knowledge obtained is passed to the Verify module, where each sub-answer is evaluated for accuracy. If a sub-answer is verified as correct, it is stored in the sub-answer pool. Should it be incorrect, a rethinking process is initiated. Subsequent sub-questions undergo a similar sequence of decomposition, retrieval, and verification. Ultimately, all sub-answers, along with the original question, are compiled by the Summarizer module to output the final answer. Solid lines denote items that are currently being processed, while dotted lines indicate items that are pending processing or are about to be addressed. The llama icon represents LLaMa-3-70B and the ChatGPT icon is GPT-3.5-turbo in this paper.}
  \label{fig3:The framework of model}
\end{figure*}
The importance of corpus cannot be overstated in the realm of external knowledge acquisition. The Wikipedia corpus released by DPR \cite{karpukhin2020dense} is widely recognized as a standard resource. However, as time progresses, this corpus faces challenges such as knowledge gaps and outdated information. Moreover, to accommodate dense retrieval requirements, the corpus is designed to fragment entity information into multiple segments, often resulting in incoherent data representation.
To address these limitations, we develop a novel Wikipedia corpus called \textbf{Indexed Wikicorpus} along with a corresponding key entity profile corpus called \textbf{Profile Wikicorpus}. Our approach differs significantly from its predecessor in prioritizing the coherence of entity information. In Indexed Wikicorpus, each entry represents a complete entity as shown in Figure \ref{fig:corpus_difference}, ensuring a more comprehensive and cohesive representation of information. Building upon this foundation, we draw inspiration from the Web version of Wikipedia to extract entity profiles. These profiles, curated by Wikipedia, encapsulate the essential information about each entity, which is saved to Profile Wikicorpus. Data analysis about the number of words and entities as shown in Table \ref{tab:corpus_diff} reveals that our new corpus contains a higher volume of entity information compared to its predecessor. Furthermore, subsequent experiments demonstrate the superior performance of our corpus over the older version. 

\section{HIERARCHICAL RETRIEVAL-AUGMENTED MODEL}

To address the challenges in retrieval-augmented generation, including the lack of deep integration between different retrieval methods and the potential introduction of noise from multi-candidate retrieval, we propose a novel framework called \textbf{HiRAG}. This framework consists of five key components: Decomposer, Summarizer, Retriever, Definer, and Filter, as illustrated in Figure \ref{fig3:The framework of model}. The detailed algorithm is presented in Algorithm \ref{alg:HiRAG_algorithm} in Appendix \ref{appedix:Algorithm}.
Specifically, the Decomposer is designed to tackle complex questions by decomposing them into smaller, more manageable sub-questions that can be easily answered. The Definer then determines whether the question can be solved. If it can, the Summarizer leverages the sub-answers to generate a response to the original question. Otherwise, the Retriever extracts relevant information related to the sub-question through a hierarchical retrieval process at both the document and chunk levels. Finally, the Filter verifies the validity of the retrieved content, generates a sub-answer, and re-evaluates the results if they are found to be inaccurate.
\subsection{Notations and Definitions}
For a multi-hop question, $x$, we can decompose it into a series of sub-questions $\mathcal{Q}$, where answers to previous sub-questions can inform the generation of subsequent sub-questions. The objective is to iteratively obtain the set of sub-questions $\mathcal{Q}=\{q_i\}_{i=1}^{|\mathcal{Q}|}$ and their corresponding set of sub-answers $\mathcal{A}=\{a_i\}_{i=1}^{|\mathcal{A}|}$, where $q_i$ represents the $i$-th decomposed sub-question and $a_i$ its corresponding sub-answer.
Ultimately, we combine all sub-answers with the original question, $x$, to get the final answer $o$.
To augment the ability of the model to answer questions, we leverage external knowledge from the Wikipedia corpus, denoted as $\mathcal{D}$, which comprises numerous sub-documents. Each sub-document, $d$, contains relevant content, including a title, $t$, and body text. For a given sub-question, we first employ a retriever to identify the most relevant sub-document, $d=\bm{Retrieval}(q,\mathcal{D})$. We then pinpoint the most relevant chunk of text, $c$, in $d$. Finally, we utilize the most relevant chunk to answer the sub-question using a LLM, yielding the sub-answer $a=\bm{LLM}(q,c)$.

\subsection{Decomposer and Summarizer }
\label{fig:leaf}
Building upon the previous work \cite{zhou2022least}, we use the Decomposer module to break down complex problems into manageable sub-questions leveraging the prompt engineering \cite{wang2024langgpt}. We develop a comprehensive prompt for the Decomposer, which includes its background, goals, constraints, workflow, examples, and initialization.
The prompt of the Decomposer module is formulated as follows:

 \begin{description}

\item\small{Prompt of Decomposer}: \\
   \colorbox{gray!20}{\parbox{0.98\linewidth}{
    \item\small{

\textbf{Background}:

- You are an expert at analyzing problems...

\textbf{Goal}:

Helping the user decompose the question and tell the user at the right time that the problem can be solved.

\textbf{Constraint}:



- You can only decompose the question, do not answer it directly...

\textbf{Workflow}:

1. Analyse the original complex question...







\textbf{Example}:

....

\textbf{Initialization}:

Now, a first simple question.}}}
\end{description}
We initialize the model turn, $mt$, to 0 which indicates the round of the current decomposition problem. For the $mt$-th iteration, the Decomposer makes a new sub-question with the previous sub-answers, $\{a_1, a_2, ...,  a_{mt-1}\}$, and original question, ${x}$. Then the $\bm{Definer}$ determines whether the original question can be solved with all known sub-answers based on the output of $\bm{Decomposer}$. Therefore, this process can be formulated as follows:
\begin{equation}
\label{decompose}
     \begin{split}
    q_{mt} & = \bm{Decomposer}(\{a_1, a_2 \dots, a_{mt-1}\}, x),\\
    \it{Judge} & = \bm{Definer}(q_{mt}),\\
    \it{mt} & = \it{mt} +1.\\
    \end{split}
\end{equation}
If the $\it{Judge}$ returns a result indicating that the original question can be solved, we send the original question, $x$, with all sub-answers before the current decomposition round to the Summarizer. The $\bm{Summarizer}$ tries to generate the output $o$ which can be formulated as follows:
\begin{equation}
    o = \bm{Summarizer}(\{a_1, a_2 \dots, a_{mt}\}, x).
\end{equation}
In Appendix \ref{appendix: model turn}, we present experimental results investigating the relationship between model performance, as measured by the EM metric, and the number of model turns $mt$. The results are summarized in Table \ref{tab:model turn}.


\subsection{Hierarchical Retriever}
In this section, we describe the design of a novel hierarchical retrieval mechanism that enhances retrieval accuracy and provides semantic clarification in cases of ambiguous semantics. Our approach employs a two-layer hierarchical retrieval process. The first layer identifies the most relevant documents within the corpus at the document level, while the second layer locates the most relevant chunks within those documents at the chunk level, resulting in a more accurate and refined retrieval process.
We begin by obtaining a decomposed question, $q$, through the Decomposer module. Next, we utilize the LLM to extract the entity name, $e$, from the question. We then leverage this entity name to identify the document with the title, $t_c$, that most closely matches the entity name among all available documents in the corpus. As this step primarily involves matching entity names with indexes, lexical matching takes precedence. Therefore, we employ sparse retrieval \cite{svore2009machine} to implement this step, capitalizing on its strengths in lexical matching.

After we get the document with the title $t_c$, we look for the corresponding information, $p^*$, in the Profile WikiCorpus and add it to the retrieval result if it exists.

\begin{equation}    
    \begin{split}
        e &= \bm{LLM}(q), \\
        t_c&=\bm{SparseRetrieval}(e, \{t_1, t_2 \dots, t_n\}),\\
        p^*&=\bm{SparseRetrieval}(t_c, \{p_1, p_2 \dots, p_m\}).
      \end{split}
      \label{eq:SparseRetrieval}
\end{equation}where $n$ represents the number of documents in the Indexd WikiCorpus and $m$ represents the number of profiles in the Profile WikiCorpus.

During the process of question decomposition and entity extraction, a notable challenge emerges. The use of LLMs for both sub-question generation and entity information extraction can lead to information loss. Specifically, when semantic loss occurs, it is often characterized by a high similarity between the extracted entity name and multiple candidate titles. To mitigate this issue, the retriever enhances the semantics by incorporating contextual information. In the context of Multi-hop QA tasks, this semantic enrichment draws upon two primary sources: the original question and the preceding answers. By leveraging this contextual information, we first select the most relevant details and then generate a new question that better captures the intended meaning. Specifically, we first select an answer, $a^*$, containing
the entity, $e$, from the set of sub-answers, $\mathcal{A}^{'}=\{a_1, a_2 \dots, a_{mt-1}\}$, which includes all sub-answers from previous $mt-1$ rounds
and utilize it as a supplement to our query. If this approach
yields no results, we instead use the original question $x$ as the supplement to guide our investigation. We 
then instruct the $\bm{LLM}$ to generate a new question $q^*$, which is subsequently retrieved using other retrieval engines.

\begin{equation}
    \begin{split}
        a^* & = \bm{Select}(\{a_1, a_2 \dots, a_{mt-1}\}, e),\\
        q^* & =\bm{LLM}(q, s).
      \end{split} \label{eq:Select}
\end{equation}
where $\bm{Select}$ is the process of finding the contextual information.

Upon obtaining the relevant sub-document $d_c$ associated with the title $t_c$, we proceed to split it into uniform chunks, $\mathcal{C} =\{c_1 , c_2, \dots,\\ c_n\}$. In this phase, we prioritize semantic matching and employ a dense retriever, Contriever \cite{izacard2021unsupervised}, to facilitate the process. The similarity score is calculated by computing the dot product of the vector representations of the question and each chunk. The chunk with the highest similarity score is then selected as the most relevant chunk, denoted as, $c_s$.

\begin{equation}
    \begin{split}
        \{c_1 , c_2 \dots, c_n \}& = \bm{Spilt}(d_c), \\
        Sim(q,c_i) &=<E(q),E(c_i)>,\ \ \forall 1 \le i \le n,\\
        c_s &= arg \max_{\mathbf{c}_i \in \mathcal{C}}\ Sim(q, c_i),\ \ \forall 1 \le i \le n,
      \end{split} \label{eq:Spilt}
\end{equation}
where $\bm{Spilt}$ is the process of dividing the documents into many chunks and n is the number of chunks.

\subsection{Filter}
Even the most advanced retrieval engines can struggle to consistently retrieve the most accurate content. To mitigate the impact of potential retrieval engine errors, we utilize a specialized filter. This filter is designed to evaluate retrieval results and refine the retrieval engine through a series of iterative adjustments when inaccuracies are detected.
Upon receiving the results from the retrieval engine $c_s$, $p^*$, and the sub-question, $q$, generated by the Decomposer, we first attempt to leverage a $\bm{LLM}$ to generate a response, $r$. The model then assesses whether the current sub-question, $q$ can be resolved based on the response, $r$.
 \begin{equation}
    \begin{split}
        r =\bm{LLM}(q, c_s, p*).
      \end{split} \label{eq:response}
\end{equation}
If the question can be solved, the Filter passes $r$ as the answer, $a$, to the sub-question to the Decomposer.
However, when faced with an unresolvable question, the Filter triggers \textbf{a two-tiered rethinking retrieval process}, encompassing both chunk-level retrieval and document-level retrieval to facilitate a more comprehensive search.

Initially, if the Filter determines that the problem cannot be resolved, it initiates a rethink at the chunk level. During this phase, the result, ${d_c}$, from the first-tier retrieval by the retriever remains unchanged. Instead, modifications are made to the second tier of the retriever, sequentially selecting chunks with higher similarity scores from the divided chunks. This approach is particularly effective when the correct entity information and corresponding references are identified, but the specific chunk required to resolve the problem is not found.
If the chunk-level rethink proves unsuccessful, the process escalates to the document level. At this level, adjustments are made to the first level of the retriever by selecting titles with higher similarity to the entity, ${e}$, from the candidate titles.
Notably, this two-tiered rethink process operates as a nested loop, where each higher-level rethink informs and drives the lower-level rethinks, ensuring a comprehensive and systematic approach to problem-solving.
Concurrently, we also address the issue of knowledge balance during the retrieval process. The rapid advancement of LLMs has led to a significant increase in the quantity and quality of their internal knowledge. As a result, a critical challenge in RAG emerges: striking a balance between the external knowledge retrieved and the internal knowledge of the model. While previous research \cite{wu2024clasheval} has focused on developing classifiers to tackle this challenge, we propose a novel and straightforward approach. By passing the number of filter rethinks as a parameter to the classifier, we observe that as the number of rethinks increases, the semantic similarity between the model and the retrieved content generally decreases.

In the context of the sub-question, $q$, if the retrieved result after the $t$-th round of rethink fails to yield an answer, we propose incorporating a probability $y$ of leveraging the internal knowledge of the model to address the question.
\begin{equation}
    \begin{split}
       y = \left(\frac{t}{m}\right)^2.
      \end{split} \label{eq:assess}
\end{equation}
where $m$ is a hyperparameter and $t$ is the round of rethink.
In our experiment, the maximum number of retrievals is set to 4, where the value of $m$ we employ is 5. We recommend that $m$ be greater than or equal to the maximum number of retrievals. This choice is the result of a trade-off. On one hand, the model should not abandon retrieval and resort to its internal knowledge too hastily, as the reliability of internal knowledge cannot be guaranteed. On the other hand, it should also ensure that when external knowledge is suboptimal and the retrieval method fails, it can attempt to utilize uncertain internal knowledge to provide an answer.




\begin{table*}[t]
\renewcommand\arraystretch{1.1}
 \caption{
    Experimental results on four datasets. Without retrieval means only using the internal knowledge of LLM while With retrieval means using the external knowledge from the retriever. HiRAG is divided into HiRAG (online) and HiRAG (local) according to whether it uses Google for retrieval after semantic supplementation. The best results are in bold.
    }
\vspace{-1em}
\resizebox{1\textwidth}{!}{
\begin{tabular}{ccccccccccccccccccccc}
\toprule[1pt]
&   & \multicolumn{4}{c}{ \textbf{HotpotQA}}                                                                              &    & \multicolumn{4}{c}{\textbf{2WikiMultihopQA}}
&   & \multicolumn{4}{c}{\textbf{MuSiQue}}
&  & \multicolumn{4}{c}{\textbf{Bamboogle}}

\\ 
\cline{3-6} \cline{8-11} \cline{12-15}  \cline{16-20} 
\multirow{-2}{*}{\textbf{Settings}}    & \multirow{-2}{*}{\textbf{Models}}  
& \textbf{EM}   & \textbf{F1}  & \textbf{Precision}   & \textbf{Recall}   &  
& \textbf{EM}   & \textbf{F1}    & \textbf{Precision}   & \textbf{Recall}  &
& \textbf{EM}   & \textbf{F1}    & \textbf{Precision}   & \textbf{Recall} &
& \textbf{EM}   & \textbf{F1}    & \textbf{Precision}   & \textbf{Recall} 
\\ 
\midrule[1pt]
& Direct & 24.00 & 31.18  & 35.51  & 29.58 &  
& 24.20 & 29.21  &  31.72   &  28.48  & 
&  2.20  &  7.15 &  9.87 &  6.20 &  
&  15.20 &  18.53 &  19.47 &  18.00 \\ 
 & CoT   & 31.80  &  43.16   &  44.14   & 45.34  &
 &  26.80   & 35.26   & 34.14     &  39.32  & 
 &  8.20 &  19.07 &  19.91 &  20.84 & 
 &  52.80 &  64.97     &  65.32 &  67.47 \\
 & CoT-SC  &  33.40   & 44.95   & 45.61  & 47.65   & 
 & 29.00 &  36.62  &  35.62  &  40.21  &
 &  10.00  &  20.33  &  21.10     &  21.89 &  
 &  52.80 &  63.55    &  63.83  &  64.30   \\
&  Self-Ask w/o Ret & 26.40   &  37.82   & 38.54   & 41.07   &  & 28.80  &  37.50   & 36.77   & 40.50 & 
&  9.20 &  19.29  &  19.36  &  22.22 &  
&  51.20 &  59.05 &  58.92 &  60.80 \\
\multirow{-5}{*}{\textbf{W/O retrieval}} 
& \textbf{HiRAG w/o Ret}   & 36.40   &  47.46  &  49.39  &  48.07    & 
& 37.20         & 46.24  & 45.08 &  49.23    & 
&  13.20   &  25.22         &  26.82          &  25.78 &  
&  \textbf{60.80} &  69.95         &  69.66         &  \textbf{71.57} \\ 
\midrule[1pt]
 &  Direct &  24.40 &  35.00  &  41.48 &  33.00   &  
 &  29.40 &  38.50 &  37.57  &  41.90  &
 &  5.20  &  10.30 &  13.91 &  9.14 & 
 &  19.20 &  27.23  &  31.60 &  25.53 \\
&  ReAct &  25.20   &  32.68  &  33.85  &  33.17  & 
&  25.20 &  30.98  &  30.81  &  32.25  &
 &  5.80 &  8.17 &  8.11 &  8.67  & 
 &  20.00  &  24.09 &  25.63 &  23.63  
\\
&  Self-Ask &  25.60 &  36.04  &  37.15 &  38.96  & 
&  35.00 &  46.42 &  45.26 &  50.99 &
 &  6.20 &  12.95 &  12.58 &  15.51 & 
 &  40.80 &  49.62  &  49.34 &  52.43
\\
&  Flare &  41.60 &  54.37 &  56.32  &  55.54 & 
&  40.60    &  52.05  &  50.34  &  57.07  &
 &  13.39   &  \textbf{26.22} &  \textbf{27.72} &  27.51  &
 &  46.34  &  57.71 &  57.50  &  58.27 \\
 & MetaRAG  &  36.80 &  48.43 &  50.41 &  49.16  &   
 &  23.15 &  30.38   &  29.79 &  32.52  &
 &  8.40  &  17.23 &  18.64 &  18.25 &  
 &  23.20  &  30.36 &  31.20 &  30.33 \\
&  \textbf{HiRAG (online)} &  40.48 &  52.51  &  53.98  & 53.70 &   
&  50.38 &  \textbf{65.30} &  \textbf{63.22} &  63.22   &
&  \textbf{14.65} &  26.03  &  26.58   & \textbf{28.15} &  
&  60.00 &  \textbf{70.34} &  \textbf{70.68} &  71.50 \\
\multirow{-7}{*}{\textbf{W/ retrieval}}    &  \textbf{HiRAG (local)}   &  \textbf{42.52} &  \textbf{54.98} &  \textbf{57.16} &  \textbf{57.16} &  &  \textbf{52.29} &  63.95        &  61.99  &  \textbf{70.27} &
&  11.11  &  21.99    &  21.89   &  25.72         &  
&  53.17   &  64.79   &  65.81    &  66.27    
\\ 
\bottomrule[1pt]
\end{tabular}

}
\label{tab:main_results}
\end{table*}

\begin{table*}[t]
 \caption{
    Result of substitution of different LLMs as the backbone, i.e. the icon of ChatGPT in Figure \ref{fig3:The framework of model}. The best results are in bold.
    }
\vspace{-1em}
\resizebox{\textwidth}{!}{
\renewcommand\arraystretch{1.1}
\begin{tabular}{ccccccccccccccccccccc}
\toprule[1pt]
& {\color[HTML]{000000} }       & \multicolumn{4}{c}{{\color[HTML]{000000} \textbf{HotpotQA}}}                    &   & \multicolumn{4}{c}{{\color[HTML]{000000} \textbf{2WikiMultihopQA}}}                                &   & \multicolumn{4}{c}{\textbf{MiSiQue}}                      &   & \multicolumn{4}{c}{\textbf{Bamgoole}}          \\ 
\cline{3-6} \cline{8-11} \cline{13-16} \cline{18-21} 
\multirow{-2}{*}{\textbf{Online or local}} & \multirow{-2}{*}{{\color[HTML]{000000} \textbf{LLMs}}} 
& \textbf{EM}    & \textbf{F1}   & \textbf{Precision}  & \textbf{Recall}     &           
& \textbf{EM}    & \textbf{F1}   & \textbf{Precision}  & \textbf{Recall}     &          
& \textbf{EM}    & \textbf{F1}   & \textbf{Precision}  & \textbf{Recall}     &  
& \textbf{EM}    & \textbf{F1}   & \textbf{Precision}  & \textbf{Recall}  \\ 
\midrule[1pt]
&  Qwen2-7B              &  41.50 & 51.65                        & 52.97                       & 53.53                       & \textbf{}               &  52.50 & 63.64 & 61.29 &  70.12 & \textbf{}               & 9.80                       & 21.87                               & 21.32                               & 25.16                    &                         & 41.60                      & 52.87                     & 54.92                    & 53.57                     \\
 &  LLaMa-3-8B              &  37.84 & 47.06                        & 49.13                       & 47.20                        & \textbf{}               &  43.60 & 54.33 & 52.64 &  58.87 & \textbf{}               & 6.80                       & 16.45                                & 16.81                                 & 17.76                      &                         & 42.86                       & 51.74                        & 51.29                     & 54.05                      \\
                                  
& {\color[HTML]{000000} LLaMa-3-70B}                   & \textbf{ 44.50} & {\textbf{ 56.10}} & {\textbf{ 58.15}} & {\color[HTML]{1F2329} 56.52} &                         & {\color[HTML]{000000} 57.00} & {\color[HTML]{000000} 68.50} & {\textbf{ 66.69}} & {\color[HTML]{000000} 74.24} &                         & 16.60                        & \textbf{ 26.27  }                                & \textbf{ 26.86}                                  & \textbf{28.18}                        &                         & 48.80                        & {\color[HTML]{000000} 62.79} & {\color[HTML]{000000} 63.19} & {\color[HTML]{000000} 63.90} \\

\multirow{-4}{*}{\textbf{Local}}    
 &  HiRAG (GPT-3.5-turbo)              &  42.52 & 54.98                        &    57.16                     &   \textbf{57.16}                      & \textbf{}               &  52.29 &  63.95 & 61.99 &  70.27 & \textbf{}               & 11.11                        & 21.99                                  & 21.89                                  & 25.72                        &                         & 53.17                        & 64.79                        & 65.81                        & 66.27                        \\

\midrule[1pt]
&  Qwen2-7B               &  32.56 & 41.51                        & 40.89                     & 44.45                       & \textbf{}               &  45.59 & 58.45 & 55.53 & 68.01 & \textbf{}               & 9.46                      & 20.57                             & 20.72                              & 22.29                    &                         &36.54                     & 50.25                     & 52.47                   & 52.56        \\
&  LLaMa-3-8B               &  37.66 & 46.28                        & 48.20                       & 46.86                        &             &  43.68 & 54.28 & 52.43 &  59.74 &              & 7.03                       & 14.86                                & 15.48                               & 15.77                      &                         &40.00                      & 49.95                        & 50.72                   & 50.83                     \\
& {\color[HTML]{000000} LLaMa3-70-B}                   & {\color[HTML]{000000} 41.12} & 52.78                        & 55.64                        & 52.88                        &                         & {\textbf{ 57.68}} & {\textbf{  70.61}} & {\color[HTML]{000000} 68.48} & {\textbf{ 77.04}} &                         & \textbf{ 15.37}                        & 25.50                                  & 26.24                                  & 26.78                        &                         & 54.76                        & 66.54      & 66.94                        & 67.33 \\ 
 \multirow{-4}{*}{\textbf{Online}}  
  & {\color[HTML]{000000} HiRAG (GPT-3.5-turbo)}                & {\color[HTML]{1F2329} 40.48} & {\color[HTML]{1F2329} 52.51} & {\color[HTML]{1F2329} 53.98} & {\color[HTML]{000000} 53.70} & {\color[HTML]{000000} } & {\color[HTML]{000000} 50.38} & {\color[HTML]{000000} 65.30} & {\color[HTML]{000000} 63.22} & {\color[HTML]{000000} 63.22} & {\color[HTML]{000000} } & {\color[HTML]{000000} 14.65} & {\color[HTML]{000000} 26.03}           & {\color[HTML]{000000} 26.58}           & {\color[HTML]{000000} 28.15} & {\color[HTML]{000000} } & {\textbf{  60.00}} & {\textbf{ 70.34}} & {\textbf{  70.68}} & {\textbf{71.50}} \\
                                        
\bottomrule[1pt]
\end{tabular}

    \label{tab:generalization}
    }
\end{table*}

\section{EXPERIMENTS SETUP}
\subsection{Datasets}
In our investigation, we conduct experiments on four datasets specifically designed for the Multi-hop question answering task, namely  \textbf{HotPotQA} \cite{yang-etal-2018-hotpotqa}, \textbf{2WikiMultiHopQA} \cite{ho2020constructing}, \textbf{MuSiQue} \cite{10.1162/tacl_a_00475}, and \textbf{Bamboogle} \cite{press2022measuring}. We only use questions and answers in the dataset, with all external knowledge retrieved.



\subsection{Baselines}
Our baselines can be divided into two categories according to whether external knowledge is retrieved: Without retrieval baselines and With retrieval baselines. The Without retrieval (\textbf{W/o retrieval}) setting includes four baselines.
\textbf{Direct} Prompting directly uses the question as a prompt for LLM to respond. For retrieval, directly use the question as the query to search. 
\textbf{CoT} Prompting \cite{wei2022chain} adds ``Let's think this question step by step'' after the initial question. 
\textbf{CoT-SC} \cite{wang2022self} consolidates its responses by answering the same question multiple times. 
\textbf{Self-Ask} \cite{press2022measuring} enables the model to decompose the problem into a series of sub-questions, allowing it to answer or retrieve relevant information for sub-questions. 
Direct and Self-Ask models belong to two working modes, so the With retrieval (\textbf{W/ retrieval}) setting includes five baselines.
\textbf{ReAct} \cite{yao2022react} is similar to self-ask, the difference is that react will extract the query to be retrieved from the sub-questions after breaking down the problem, while Self-ask will directly use the sub-questions as the query.
\textbf{Flare} \cite{jiang2023active} uses a confidence-based strategy when generating questions. It generates questions about the less-confident parts of the generated content and retrieves them. 
\textbf{MetaRAG} \cite{zhou2024metacognitive} gains insights from metacognition, which can identify logical errors in reasoning and use the three-step metacognitive regulation pipeline to identify and repair deficiencies in initial cognitive responses.

\subsection{Implementation Details}
We use \texttt{GPT-3.5-turbo} as the backend LLM for the most part while using \texttt{LLaMa-3-70B} for the Decomposer. For baselines, we use the family of \texttt{GPT-3.5} as the backend LLM. For online retrieval, we use Wikipedia and Google through the Wikipedia API and Serper API. For local retrieval, we use Contriever-MSMARCO \cite{izacard2021unsupervised} and elastic with the BM25 algorithm \cite{robertson2009probabilistic} as dense retriever and sparse retriever. The default maximum number of retrievals is 4. For HiRAG, we use Indexed Wikicorpus and Profile Wikicorpus for retrieval while we use the DPR corpus \cite{karpukhin2020dense} for retrieval in baselines. Following \cite{zhou2024metacognitive}, we employ four evaluation metrics to assess the performance of our model. At the answer level, we use Exact Match (EM) to determine whether the predicted answer aligns perfectly with the reference answer. Additionally, at the token level, we adopt a more fine-grained approach, which includes token-level F1 score, precision, and recall. Following MetaRAG \cite{zhou2024metacognitive}, for cost considerations, we sub-sample 500 questions
from the validation set of HotPotQA, 2WikiMultiHopQA, and MuSiQue for experiments while using the full set of Bamboogle for experiments as the quantity of Bamboogle is small.

\begin{table*}[h]
 \caption{
    Ablation experimental results for the Retriever module.
    }
\vspace{-1em}
\resizebox{\textwidth}{!}{
\renewcommand\arraystretch{1.1}
\begin{tabular}{cccccccccccccccccccc}
\toprule[1pt]
& \multicolumn{4}{c}{{\color[HTML]{000000} \textbf{HotpotQA}}}                                                                       &           & \multicolumn{4}{c}{{\color[HTML]{000000} \textbf{2WikiMultihopQA}}}                                                                &           & \multicolumn{4}{c}{\textbf{MuSiQue}}         &  & \multicolumn{4}{c}{\textbf{Bamgoole}}                                                                         \\ \cline{2-5} \cline{7-10} \cline{12-15} \cline{17-20} 
\multirow{-2}{*}{\textbf{Online or local}}            
& \textbf{EM}    & \textbf{F1}   & \textbf{Precision}  & \textbf{Recall}     &           
& \textbf{EM}    & \textbf{F1}   & \textbf{Precision}  & \textbf{Recall}     &          
& \textbf{EM}    & \textbf{F1}   & \textbf{Precision}  & \textbf{Recall}     &  
& \textbf{EM}    & \textbf{F1}   & \textbf{Precision}  & \textbf{Recall}  \\ 
\midrule[1pt]
\textbf{HiRAG}                                          & 42.52 & 54.98                        & 57.16                        & 57.16                        & \textbf{} &  52.29 &  63.95 &  61.99 &  70.27 & \textbf{} & 11.11 & 21.99 & 21.89     & 25.72  &  & 53.17 & 64.79                        & 65.81                        & 66.27                        \\
\textbf{HiRAG (w/o chunk)}            & 41.85 & 53.01                        & 55.16                        & 54.17                       & \textbf{} &  49.04 &  62.69 &  61.00 &  67.85 & \textbf{} & 6.02 & 15.92 & 15.23     & 18.95  &  & 38.40 & 58.58                        & 57.46                        & 62.50                       \\ 
\textbf{HiRAG (w/o document)}                                & 39.67 & 52.30                      & 54.70                       & 52.92                       & \textbf{} &  48.68 &  61.52 &  59.63 &  67.17 & \textbf{} & 9.45 & 18.37 & 21.85    & 20.37  &  & 48.36 & 58.72                       & 59.56                       & 59.70                      \\

\textbf{HiRAG (w/o chunk + document)}          & 39.57 & 50.36                     & 50.52                     & 51.04                       & \textbf{} &  47.50 &  53.87 &  52.62 & 56.70 & \textbf{} & 5.51 & 13.97 & 13.95    & 15.35 &  & 42.53 & 53.37                     & 54.79                      & 54.02                    \\
\textbf{HiRAG (w/o Profile WikiCorpus)}          & 40.98 & 52.84                   & 54.51                 & 53.96                       & \textbf{} &  50.67 &  65.04 &  63.09 & 71.09 & \textbf{} & 10.81 & 21.54 & 20.81  & 25.68 &  & 51.61 &63.39                     & 65.59                     & 64.78                   \\
\bottomrule[1pt]
\end{tabular}
}

    \label{tab: Ablation_results}

\end{table*}

\begin{table*}[h]
\caption{
    Corpus experiments based on Flare, such as the corpus generated by DPR and Indexed Wikicorpus. 
    }
\vspace{-1em}
\renewcommand\arraystretch{1.1}
\resizebox{\textwidth}{!}{
\begin{tabular}{cccccccccccccccccccc}
\toprule[1pt]
\multirow{2}{*}{\textbf{Corpus}}   
& \multicolumn{4}{c}{\textbf{HotpotQA}}                      
&  & \multicolumn{4}{c}{\textbf{2WikiMultihopQA}}            &  & \multicolumn{4}{c}{\textbf{MuSiQue}}                    &  & \multicolumn{4}{c}{\textbf{Bamboogle}}                  \\ \cline{2-5} \cline{7-10} \cline{12-15} \cline{17-20} 
& \textbf{EM}    & \textbf{F1}  & \textbf{Precision}    & \textbf{Recall}    &                      
& \textbf{EM}    & \textbf{F1}  & \textbf{Precision}    & \textbf{Recall}   &
& \textbf{EM}    & \textbf{F1}  & \textbf{Precision}    & \textbf{Recall} &
& \textbf{EM}    & \textbf{F1}  & \textbf{Precision}    & \textbf{Recall} \\
\midrule[1pt]

\multicolumn{1}{c}{\textbf{Old corpus (DPR)}}      & 41.60          & 54.37          & 56.32          & 55.54          & \multicolumn{1}{c}{} & 40.60          & 52.05          & 50.34          & 57.07          &                      & 13.39          & \textbf{26.22} & \textbf{27.72} & \textbf{27.51} &                      & 46.34          & 57.71          & 57.50          & 58.27          \\
\multicolumn{1}{c}{\textbf{Indexed Wikicorpus}} & \textbf{43.00} & \textbf{55.06} & \textbf{56.95} & \textbf{55.65} & \multicolumn{1}{c}{} & \textbf{50.00} & \textbf{62.27} & \textbf{60.30} & \textbf{68.15} &                      & \textbf{13.60} & 25.62          & 26.55          & 26.58          &                      & \textbf{49.60} & \textbf{60.20} & \textbf{60.28} & \textbf{61.13} \\ \bottomrule[1pt]
\end{tabular}
    \label{tab:corpus}
    }
\end{table*}
\section{RESULTS AND ANALYSIS}
\subsection{Main Results}


The main results are shown in Table \ref{tab:main_results} and reveal the following notable findings.
(1) Our proposed framework, HiRAG, exhibits superior performance across multiple evaluation metrics, outperforming state-of-the-art methods on three out of four datasets. Additionally, it demonstrates improvements in several metrics on the remaining dataset. The key advantage of our approach is its emphasis on the retrieval process, which consistently produces high-quality results. This highlights the crucial role of the retrieval component in achieving exceptional outcomes.
(2) In the Without Retrieval setting, HiRAG significantly outperforms the baselines, showcasing its effectiveness in question decomposition and answer summarization. This is due to the framework's design, which enables autonomous sub-question generation and termination of the loop when the original problem is resolvable. Our approach also differs from self-ask in its segregation of subtasks and implementation of each subtask through a separate prompt, resulting in superior performance.
(3) A comparative analysis of results across datasets reveals that HiRAG achieves the most significant breakthrough in 2WikiMultiHopQA, with a notable improvement of over 12\% in the EM index compared to the state-of-the-art method. This is primarily attributed to the fact that most external knowledge required by 2WikiMultiHopQA can be retrieved from Wikipedia, and the question format in this dataset is relatively standardized. However, the MuSiQue dataset poses a greater challenge due to the complexity of the questions and the inability to directly retrieve required knowledge from Wikipedia. This complexity affects our framework's ability to evaluate retrieval results, diminishing the effectiveness of the response.

\subsection{Generalization}
To assess the generalizability of HiRAG, we conduct a series of experiments where we substitute different base models, with the results presented in Table \ref{tab:generalization}. Specifically, we evaluate the performance of HiRAG when paired with several prominent base models, including LLaMA-3-70B, LLaMA-3-8B, and Qwen2-7B. The results demonstrate that our method exhibits robustness and effectiveness across different base models, showcasing its adaptability to various model architectures. When paired with LLaMA-3-70B, our approach achieves state-of-the-art performance on most datasets, underscoring its ability to enhance the performance of strong base models and highlighting its potential for widespread applicability. 
\textit{Notably, to ensure a fair comparison with existing SOTA models, like Meta-RAG, which is built on GPT-3.5-turbo, we use the same base model in our experiments, even though LLaMA-70B has shown better performance.} This allows for a more direct and meaningful comparison of our approach with prior work.

\subsection{Ablation Experiments}
We conduct ablation experiments to evaluate the contributions of our Retriever module, focusing on the hierarchical retrieval approach. By removing the processing during rethinking at the chunk level and document level, we investigate the impact of these components on the overall performance. The results, presented in Table \ref{tab: Ablation_results}, demonstrate the effectiveness of our approach.
Our method is designed to improve the accuracy of retrieved results through chunk level and document level rethink. To isolate the impact of each level, we perform ablation experiments by eliminating one level of rethink at a time. Specifically, we remove chunk level rethink by considering only the highest-scoring chunk within a document and remove document level rethink by retrieving only the highest-scoring document. The results show that eliminating either level of rethink leads to a decrease in performance, confirming the importance of both chunk level and document level rethink in our hierarchical retrieval approach.
Furthermore, we investigate the influence of knowledge incorporated in the Profile WikiCorpus. Our findings indicate that removing this knowledge component leads to a decrease in the performance of HiRAG.

\subsection{Indexed Wikicorpus}

We conduct experiments under the FLARE framework to assess the effectiveness of our corpus in improving model performance. As shown in Table \ref{tab:corpus}, replacing the external knowledge source with our corpus leads to notable advancements in Exact Match (EM) scores across four datasets, with three datasets experiencing improvements in all evaluated metrics. The comprehensive coverage and meticulous entity name segmentation in our corpus are key factors contributing to these improvements, enabling more efficient retrieval and utilization of relevant information.

\subsection{Retriever Module as Plug-in}
We extract retriever and filter modules HiRAG,  pairing them with Indexed Wikicorpus and Profile Wikicorpus to create a novel intelligent retriever. This approach leverages LLMs to evaluate and refine retrieval results, differing from traditional retrievers. We conduct experiments on standardized RAG tasks for \textit{single-hop questions using gold decomposed question-answer pairs} from 2WikiMultiHopQA and MuSiQue datasets. We compare the performance of five methods: (1) Direct Answering, (2) Sparse Retrieval with Elastic and BM25, (3) Dense Retrieval with Contriever-MSMARCO, (4) HiRAG (Online), and (5) HiRAG (Local).
The results in Table \ref{tab:plugin} clearly demonstrate the superiority of our retrieval engine over traditional approaches, with significant gains observed even in local retrieval scenarios. Specifically, we report a maximum improvement of over 9\% in EM metric, outperforming current mainstream retrieval engines. Moreover, our engine exhibits robust and comprehensive improvements across all datasets and evaluation metrics, underscoring the effectiveness of our hierarchical retrieval method. The ability of our engine to serve as a plug-in, augmenting the performance of other methods, makes it a valuable tool for achieving state-of-the-art results in a variety of applications.
\begin{table}[h]
 \caption{
    Comparison of HiRAG as a Plug-in with other retrieval engines. 
    }
\begin{adjustbox}{width=0.48\textwidth}
\renewcommand\arraystretch{1}
\begin{tabular}{cccccc}
\toprule[1pt]
                                  & {\color[HTML]{000000} }                                  & {\color[HTML]{000000} }                     &                                       &                                       &                                       \\
\multirow{-2}{*}{\textbf{Dataset}}         & \multirow{-2}{*}{{\color[HTML]{000000} \textbf{Retrieval method}}} & \multirow{-2}{*}{{\color[HTML]{000000} \textbf{EM}}} & \multirow{-2}{*}{\textbf{F1}}                  & \multirow{-2}{*}{\textbf{Precision}}           & \multirow{-2}{*}{\textbf{Recall}}              
\\ \midrule[1pt]
                                  & {\color[HTML]{000000} Direct}                            & {\color[HTML]{1F2329} 10.93}                & {\color[HTML]{1F2329} 16.59}          & {\color[HTML]{1F2329} 17.04}          & {\color[HTML]{1F2329} 17.42}          \\
                                  & {\color[HTML]{000000} Sparse retrival}                   & {\color[HTML]{1F2329} 49.28}                & {\color[HTML]{1F2329} 54.91}          & {\color[HTML]{1F2329} 55.69}          & {\color[HTML]{1F2329} 55.99}          \\
                                  & {\color[HTML]{000000} Dense retrieval}                    & {\color[HTML]{1F2329} 46.06}                & {\color[HTML]{1F2329} 54.10}          & {\color[HTML]{1F2329} 55.85}          & {\color[HTML]{1F2329} 54.96}          \\
                                  & HiRAG (online)                                              & {\color[HTML]{1F2329} \textbf{57.46}}       & {\color[HTML]{1F2329} \textbf{76.33}} & {\color[HTML]{1F2329} \textbf{75.75}} & {\color[HTML]{1F2329} \textbf{79.66}} \\
                                 
\multirow{-5}{*}{\textbf{2WikiMultihopQA}}  & {\color[HTML]{000000} HiRAG (local)}                        & {\color[HTML]{1F2329} 55.22}                & {\color[HTML]{1F2329} 73.77}          & {\color[HTML]{1F2329} 73.00}          & {\color[HTML]{1F2329} 77.23}          \\
\midrule[1pt]
& {\color[HTML]{000000} Direct}                            & {\color[HTML]{1F2329} 9.61}                 & {\color[HTML]{1F2329} 12.29}          & {\color[HTML]{1F2329} 12.62}          & {\color[HTML]{1F2329} 12.85}          \\
                                  & {\color[HTML]{000000} Sparse retrieval}                   & {\color[HTML]{1F2329} 26.60}                & {\color[HTML]{1F2329} 33.77}          & {\color[HTML]{1F2329} 34.92}          & {\color[HTML]{1F2329} 34.67}          \\
                                  & {\color[HTML]{000000} Dense retrieval}                    & {\color[HTML]{1F2329} 29.99}                & {\color[HTML]{1F2329} 38.95}          & {\color[HTML]{1F2329} 39.97}          & {\color[HTML]{1F2329} 40.64}          \\
                                  & {\color[HTML]{000000} HiRAG (online)}                       & {\color[HTML]{1F2329} \textbf{52.10}}       & {\color[HTML]{1F2329} \textbf{73.01}} & {\color[HTML]{1F2329} \textbf{74.09}} & {\color[HTML]{1F2329} \textbf{76.82}} \\ 
                                 
\multirow{-5}{*}{\textbf{MuSiQue}}          
 & {\color[HTML]{000000} HiRAG (local)}                        & {\color[HTML]{1F2329} 32.13}                & {\color[HTML]{1F2329} 44.06}          & {\color[HTML]{1F2329} 42.48}          & {\color[HTML]{1F2329} 48.14}          \\
\bottomrule[1pt]
\end{tabular}
\end{adjustbox}
    \label{tab:plugin}
\end{table}


\subsection{Case Study}

We provide a case study in Figure \ref{fig:case study} of Appendix \ref{appendix:case_study} to demonstrate the workings of our framework. We gradually generate sub-questions, retrieve sub-questions, evaluate the retrieval results, and answer the sub-question. When the retrieval results are not ideal, we use rethink to find a more satisfactory result. Finally, we use the answers to all sub-questions to get the answer to the original question.

\section{CONCLUSION}
We present a novel framework for multi-hop question answering, addressing the challenges of outdated and insufficient knowledge, context window limitations, and accuracy-quantity trade-offs. Our proposed framework, HiRAG, incorporates a hierarchical retrieval strategy, single-candidate retrieval, and a rethink mechanism to improve the efficiency and effectiveness of knowledge retrieval. The experimental results demonstrate that HiRAG outperforms state-of-the-art models on multiple datasets, confirming the effectiveness of our approach. Additionally, our newly constructed corpora, Indexed Wikicorpus which is shown to be more comprehensive and logically organized, and Profile Wikicorpus. Our contributions provide a significant step forward in improving the performance of multi-hop QA tasks, and we believe that our framework and corpora will be valuable resources for future research in this area. In the future, we hope to expand our research on the retrieval-verify-rethink pipeline to achieve more fine-grained and accurate retrieval.

\begin{acks}
This research is supported by the National Natural Science Foundation of China (No.62272092, No.62172086, and
No.62106039).
\end{acks}

\bibliographystyle{ACM-Reference-Format}
\bibliography{sample-base}

\appendix
\section{APPENDIX}

\label{appedix:Algorithm}
\subsection{Experiment about Model Turn}
\label{appendix: model turn}
We investigate the relationship between the number of model turn and the EM score, focusing exclusively on instances where the frequency exceeded 1\%. As presented in Table \ref{tab:model turn}, our results show that the model achieves relatively high EM when the number of jumps is 2 and 5. This can be attributed to the fact that most of our test datasets consist of two-hop problems. When the model successfully decomposes the problem into sub-problems that align with the dataset, it yields better results. Additionally, since we set the maximum number of decomposed sub-problems to 5, the model performs well when it breaks down the problem into more detailed and granular components.

\begin{table}[h]
 \caption{
    Analyze the relationship between the number of model turn and EM. 
    }
    \vspace{-1em}
\resizebox{0.48\textwidth}{!}{
\renewcommand\arraystretch{1}
\begin{tabular}{cccccc}
\toprule[1pt]
\textbf{Models}     &\textbf{1}      & \textbf{2}              & \textbf{3}     & \textbf{4}      & \textbf{5}       \\ \midrule[1pt]
\textbf{HotpotQA}          & 47.37  & \textbf{49.76} & 33.59 & 27.66 & 44.44  \\
\textbf{2WikiMultihopQA} & 14.27  & 45.25 & 33.93 & 72.26 & \textbf{57.14}  \\
\textbf{MuSiQue}          & $\sim$ & \textbf{24.84} & 17.93 & 5.30  & $\sim$ \\
\textbf{Bamboogle}          & $\sim$ & 64.04 & 50.00 & 42.86 & \textbf{66.67}  \\ \bottomrule[1pt]
\end{tabular}
}
\label{tab:model turn}
\end{table}







\begin{figure*}[h]
    \centering
    \includegraphics[width=0.95\textwidth]{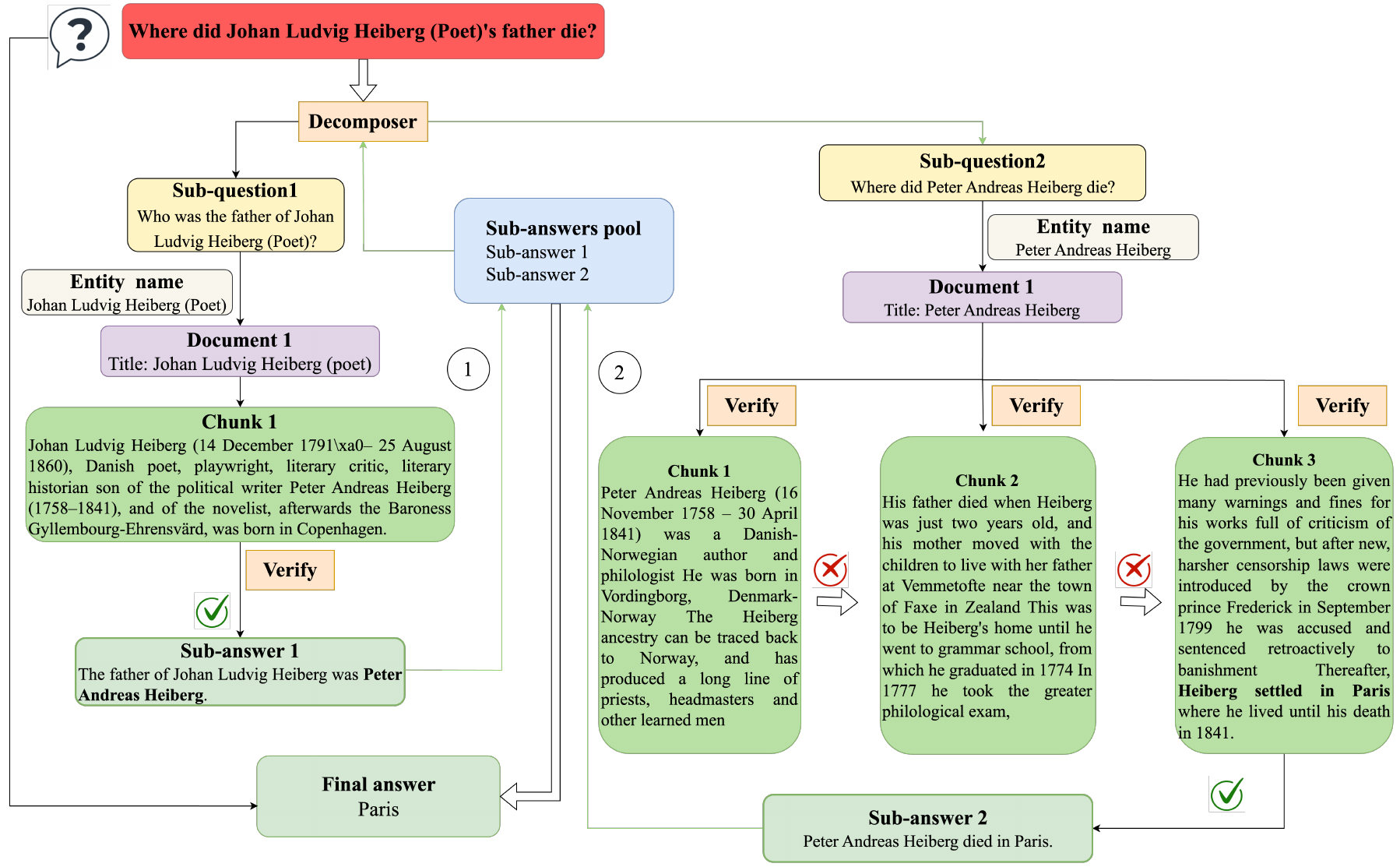}
  \caption{Case study of HiRAG process. }
  \label{fig:case study}
\end{figure*}
\subsection{Algorithm of HiRAG}

For clarity, we outline the workflow of the HiRAG framework in Algorithm \ref{alg:HiRAG_algorithm}, providing a step-by-step illustration of our method. This algorithm contains the Main function, Decompose function, RewriteAndAnswerQuestion function and Retrieval function. The Decompose function and the main function complete the decomposition and final summary of the problem, the Retrieval function describes the hierarchical Retrieval and the process of the Filter, and the RewriteAndAnswerQuestion function describes our solution when the semantics are found to be incomplete.
\SetAlgoLined
\SetAlgoVlined
\begin{algorithm*}[h]
 \caption{Description of the process of HiRAG.}
    \label{alg:HiRAG_algorithm}
	\SetKwInOut{Input}{Input}
	\SetKwInOut{Output}{Output}
        \SetKwInOut{Initialize}{Initialize}
	\SetKwFunction{NotEqual}{NotEqual}
	\SetKwFunction{add}{add}
	\SetKwFunction{TemplateVerify}{TemplateVerify}
	\SetKwFunction{TemplateComlete}{TemplateComlete}
	\SetKwFunction{ChainGenerate}{ChainGenerate}
	\SetKwFunction{Tracing}{Tracing}
	\SetKwFunction{LLM}{LLM}
	\SetKwFunction{AddChild}{AddChild}
        \SetKwFunction{Main}{Main}
	\Input{Multi-hop question: \ $x$;\ Large Language Model: LLM; }
        \Initialize{Decompose rounds: $w $;\  Corpus: $D = \{d_1, d_2 \dots, d_m$\};\  Previous answers: $ A= \{a_1, a_2 \dots, a_{w-1}\}$; \ Previous questions: $ Q= \{q_1, q_2 \dots ,q_{w-1}\}$;\ Rethink rounds: $t $;\ Threshold: $th_1$, $th_2$, $th_3$.}
    \Output{Final answer: $o$.}  
	
    \SetKwFunction{DuplicateQuery}{DuplicateQuery}

    \SetKwFunction{PromptForVerify}{PromptForVerify}
    \SetKwFunction{PromptForComplete}{PromptForComplete}
    
    \SetKwFunction{Decompose}{Decompose}
    \SetKwFunction{SparseRetrieval}{SparseRetrieval}
    \SetKwFunction{DenseRetrieval}{DenseRetrieval}
    \SetKwProg{Fn}{Function}{:}{}
    \Fn{\Decompose{$x$, $w$, $A$}}{
        $q =$ \LLM {$x$, $w$, $A$};
        \textit{//Use the original question and previous sub-answers to get new sub-question}\\
        \textbf{return} $q $  \;
    }
    \SetKwFunction{Google}{Google}
    \SetKwFunction{RewriteQuestion}{RewriteAndAnswerQuestion}
    \SetKwProg{Fn}{Function}{:}{}
    \Fn{\RewriteQuestion{$q, A, x, D, e$}}{
         \textit{//There is the case of semantic incompleteness}\\
         $IncFlag$ =Flase;\\
         \ForEach{$a_i$ in $A$}{
            \If{$e$ in $a_i$}
                {$q^*$ =\LLM{$q,a_i$};\textit{//rewrite the question}\\
                 $IncFlag$ =True;

            }
         }
         \If{$IncFlag==Flase$}{
         \textit{//previous sub-answers don't contain question's entity}\\
         $q^*$ =\LLM{$q,x$};\textit{//rewrite the question}\\
         
         }
         $c_s$=\Google{$q^*$}; ~
         $a_w$=\LLM{$c_s, q$};\\
         \textbf{return} $a_w $  \;
    }
    \SetKwFunction{Retrieval}{Retrieval}
    \SetKwProg{Fn}{Function}{:}{}

    \Fn{\Retrieval{$q$, $D$, $x$, $A$}}{
	$e =$\LLM{$q$}; ~ $t =0$; ~ \textit{//get the entity name}\\

        \While{True}{
            $t =t+1$;\\
            $d_s$=\SparseRetrieval{$e, D, t$};\textit{//get the document }\\
            \If{$len(d_s)>th_1$}{ 
            \textit{//Multiple entities have high similarity and there is semantic incompleteness }

                \textbf{return} \RewriteQuestion{$q, A, x, D, e$}; }
                 $c_s$=\DenseRetrieval{$d_s, q, t$};\textit{//get the chunk}\\
                $CanSolved$=\LLM{$c_s, q$};\textit{//with the retrieval chunk can LLM solve the question }\\
            \If{$CanSolved == "yes"$}{
                    \textit{// LLM can solve the question}\\
                    $a_w$=\LLM{$c_s, q$};\textit{//get the answer for the question? }\\
                    \textbf{return} $a_w $  \;}
                \Else{    
                
                \If{$t>=th_2$ and ($t \bmod  th_2$ == 0)}
                {
                     \textit{//the chunk level rethink fail, go to the document level rethink}\\
                     $d_s$=\SparseRetrieval{$e, D, t$};\textit{//get other document }
                }
                \If{$t>=th_3$}
                {
                     \textit{//Too many attempts, empty result returned}\\
                      \textbf{return} ''  \;
                }
                
                 \textit{//go to the chunk level rethink} 
            }
        }


                
                
                
      }
    \SetKwFunction{Main}{Main}
    \SetKwProg{Fn}{Function}{:}{}
    \Fn{\Main{$x$, $w$, $D$, $A$, $Q$}}{
        $q =$\Decompose{$x$, $w$, $A$};\\
        \While{not ($q ==$ ``That's enough'' or $q$ in $Q$)}{

    $a_w =$\Retrieval{$q$, $D$, $x$, $A$};\quad \textit{// Get the answer.}\\
    $w=w+1$;\\
    $q =$\Decompose{$x$, $w$, $A$};}
    $Output =$\LLM{$x, A$};\textit{// Get the output for original question using all sub-answers.}\\
    \textbf{return}  $Output$; 
    }
   
\end{algorithm*}

\subsection{Case Study}
We illustrate the effectiveness of the HiRAG framework through a case study on a multi-hop question, as depicted in Figure \ref{fig:case study}. Specifically, the original question presented in the figure is decomposed into two sub-questions, which are then retrieved and verified in a layered manner. Sub-questions that fail verification are re-evaluated through a rethink process. Notably, the second sub-question in the figure yields the correct result after multiple chunk-level rethinks. Ultimately, we combine the sub-answers to form the final answer to the original question.
\label{appendix:case_study}

\end{document}